\documentclass[10pt,twocolumn,letterpaper]{article}

\usepackage{cvpr}              
\usepackage{pifont}
\usepackage{booktabs}
\usepackage{xcolor}
\usepackage[accsupp]{axessibility}  

\definecolor{cvprblue}{rgb}{0.21,0.49,0.74}
\usepackage[pagebackref,breaklinks,colorlinks,allcolors=cvprblue]{hyperref}

\def\paperID{18434} 
\def\confName{CVPR}
\def\confYear{2025}

\title{GUI-Xplore: Empowering Generalizable GUI Agents with One Exploration}

\author{
Yuchen Sun$^{1}$ \quad Shanhui Zhao$^{2}$ \quad Tao Yu$^{2}$ \quad Hao Wen$^{2}$
\\\quad Samith Va$^{1}$ \quad Mengwei Xu$^{5}$ \quad Yuanchun Li$^{24*}$ \quad Chongyang Zhang$^{13}$\thanks{Corresponding authors.} \\
$^{1}$ School of Information Science and Electronic Engineering, Shanghai Jiao Tong University \\
$^{2}$ Institute for AI Industry Research (AIR), Tsinghua University\\
$^{3}$ MoE Key Lab of Artificial Intelligence, AI Institute, Shanghai Jiao Tong University\\
$^{4}$ Beijing Academy of Artificial Intelligence (BAAI)\\
$^{5}$ Beijing University of Posts and Telecommunications\\
}

\begin{document}
\maketitle
\begin{abstract}
GUI agents hold significant potential to enhance the experience and efficiency of human-device interaction. However, current methods face challenges in generalizing across applications (apps) and tasks, primarily due to two fundamental limitations in existing datasets. First, these datasets overlook developer-induced structural variations among apps, limiting the transferability of knowledge across diverse software environments. Second, many of them focus solely on navigation tasks, which restricts their capacity to represent comprehensive software architectures and complex user interactions.  To address these challenges, we introduce GUI-Xplore, a dataset meticulously designed to enhance cross-application and cross-task generalization via an exploration-and-reasoning framework. GUI-Xplore integrates pre-recorded exploration videos providing contextual insights, alongside five hierarchically structured downstream tasks designed to comprehensively evaluate GUI agent capabilities. To fully exploit GUI-Xplore's unique features, we propose Xplore-Agent, a GUI agent framework that combines Action-aware GUI Modeling with Graph-Guided Environment Reasoning. Further experiments indicate that Xplore-Agent achieves a 10\% improvement over existing methods in unfamiliar environments, yet there remains significant potential for further enhancement towards truly generalizable GUI agents. \footnote{Dataset and Code are available at \url{https://github.com/921112343/GUI-Xplore}.}

\end{abstract}    
\section{Introduction}
\label{sec:intro}
\begin{figure*}[t]
  \setlength{\abovecaptionskip}{0pt}
  \centering
  \includegraphics[width=1\linewidth]{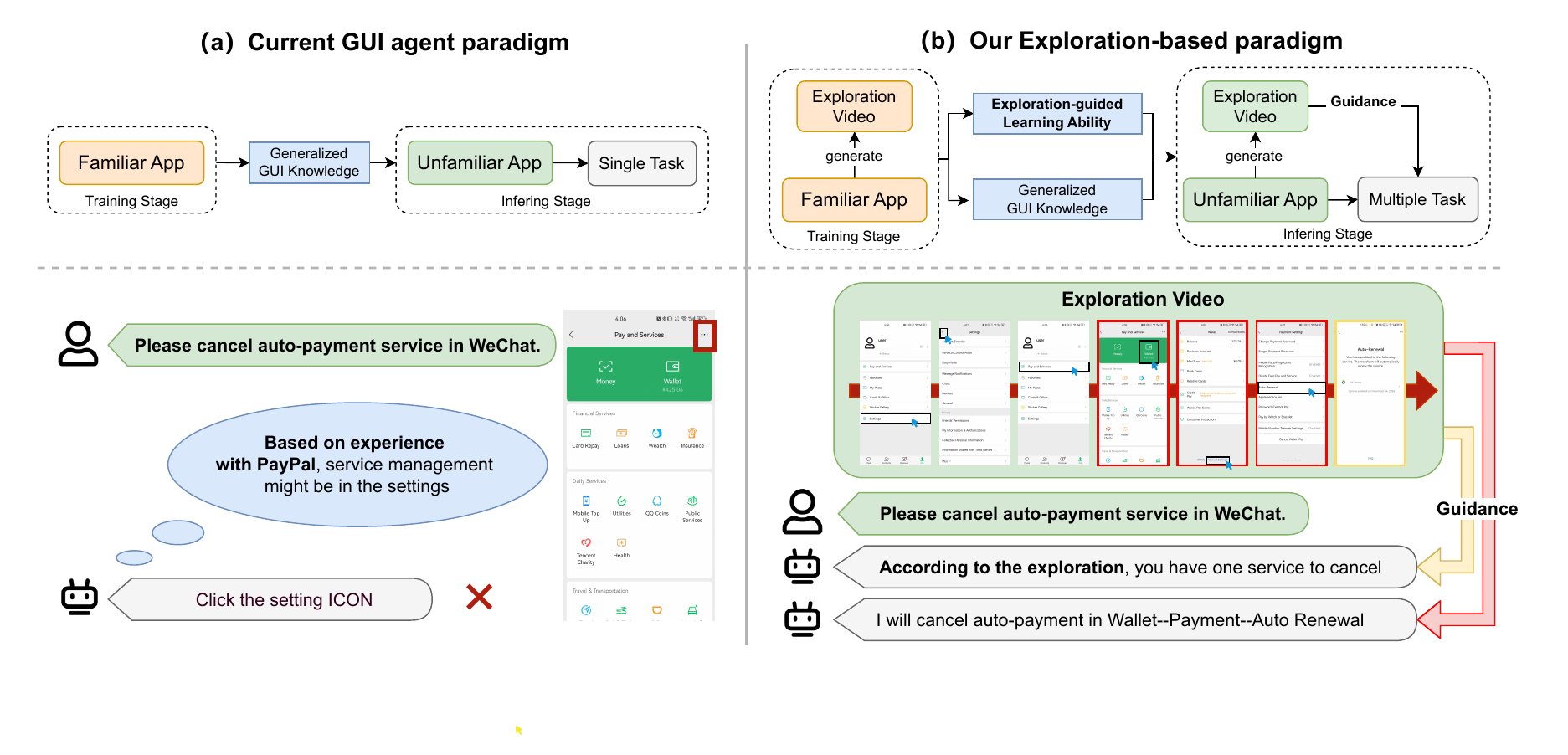}
  \vspace{-8mm} 
  \caption{Comparison between the current GUI agent paradigm and our exploration-based paradigm. (a) The current paradigm only learns generalized GUI knowledge during the training stage, lacking app-specific knowledge for inference in unfamiliar apps. For example, experience with PayPal can not translate to guidance for operating WeChat. (b) Our exploration-based paradigm provides exploration videos for each app, offering rich information of the entire app, that enable the model to learn both generalized GUI knowledge and exploration-guided learning ability. In this example, by equipping the GUI agent with knowledge from the exploration video, it can not only identify proper operation sequence for a given task, but also provide additional information according to the exploration.}
  \label{fig:intro}
  \vspace{-3mm}
\end{figure*}

With the growing integration of personal computers and smartphones, Graphical User Interfaces (GUIs) have become the predominant medium for human-device interaction \cite{Backlinko2024Smartphone}. GUI agents, envisioned as virtual personal assistants, not only simplify software operations but also offer personalized feedback, thereby holding the potential to transform the interaction paradigm. However, achieving a highly generalizable GUI agent that operates seamlessly across varied apps and tasks remains challenging, due to the limitations of existing training frameworks and datasets.

In recent years, datasets for GUI agents have expanded in scope, aiming to benchmark models' effectiveness in executing user instructions. Some datasets \cite{deng2024mind2web,rawles2024androidinthewild, chen2024guicourse, cheng2024seeclick} leverage extensive real-world interaction data to enhance task planning and execution. Others simulate interactive environments \cite{zhou2023webarena, trivedi2024appworld, zhang2024llamatouch, erdogan2024tinyagent} with structured action spaces enhance operation abilities in specific environments. Despite these advancements, critical limitations remain in supporting generalizable GUI agents. \textbf{(1) Limited Cross-App ability:} Many datasets lack sufficient app-specific knowledge, which is essential for agents to function effectively in unfamiliar apps. As shown in Fig \ref{fig:intro}(a), Developer-driven structural variations (like design choices or architectural differences) across apps lead to distinct interaction logic that hinder effective knowledge transfer, thereby limiting agent adaptability on unfamiliar environments. \textbf{(2) Limited Cross-task ability:} Most of current datasets focus primarily on basic automation tasks, neglecting the broader spectrum of user intentions and complex interactions that are essential for generating comprehensive GUI agent.

Inspired by users' exploration-and-reasoning strategies when using unfamiliar software, we propose GUI-Xplore dataset to enhance model generalization across apps and tasks through the following key innovations:

\textbf{1. Cross-App Generalization via Exploration Videos}: 
To address the generalization limitations of existing datasets, GUI-Xplore incorporates pre-recorded exploration videos for each app. These videos capture critical information, including GUI element components, navigation sequences, and user interaction logic, which serve as inductive priors for GUI agents. As illustrated in Fig\ref{fig:intro}(b), the agent, enhanced with exploration-guided learning ability, effectively utilizes this prior knowledge to facilitate efficient knowledge transfer, enabling quicker adaptation to novel software environments.

\textbf{2. Cross-Task Versatility Beyond Basic Navigation}: 
Moving beyond a narrow focus on automation, GUI-Xplore provides diverse task annotations covering Page Analysis, Application Usage, Application Overview, Action Recall, and Action Sequence Verification. This comprehensive task diversity reflects real-world GUI usages, setting a benchmark for evaluating multifunctional GUI agents.

To fully exploit GUI-Xplore’s unique features, we further propose a simple baseline framework: Xplore-Agent, which leverages exploration videos to support stable agent performance in unfamiliar apps. Given the unstructured nature of exploration videos, we employ Action-aware keyframe extraction to identify keyframes based on GUI actions. Furthermore, we construct a GUI transition graph to model the complex relationships between GUI screens, which provides prompts for the LLM to perform five downstream tasks. Despite its simplicity, our Xplore-Agent remain effective. Experimental results indicate that Xplore-Agent achieves a 10\% improvement over state-of-the-art (SOTA) methods in previously unseen apps, highlighting the utility of exploration-based priors for cross-App generalization. Additionally, benchmarking SOTA methods across GUI-Xplore’s five hierarchical downstream tasks revealed significant performance discrepancies, indicating that current methods still fall short of achieving a fully generalizable GUI agent. This analysis underscores the value of GUI-Xplore in advancing the development of versatile and generalizable GUI agents.

Our contributions are threefold as follows:

\textbf{1. A Challenging Task for GUI Agent}: We propose the app exploration task, which focuses on the ability of the GUI agent in cross-app and cross-task scenarios.

\textbf{2. The GUI-Xplore Dataset}: We present a large-scale dataset supporting the App Exploration task, comprising exploration videos from 312 apps, with over 32569 Q\&A pairs of multi-level downstream tasks. This dataset establishes a robust foundation for developing GUI agents capable of managing diverse and complex user interactions.

\textbf{3. A Baseline Framework}: We propose a two-stage baseline tailored for GUI-Xplore, enabling a comprehensive understanding of app environments. Xplore-Agent achieves a 10\% improvement over existing methods in unfamiliar apps, underscoring the efficacy of exploration-based priors for cross-app generalization.

\section{Related Works}
\label{sec:related}

\subsection{GUI Benchmarks}

Existing benchmarks for generalized GUI agents can be classified into two categories based on their training paradigm: pre-trained frameworks and interactive environment frameworks. The former provides extensive interaction data from real app environments \cite{deng2024mind2web, rawles2024androidinthewild, chen2024guicourse, cheng2024seeclick}, enabling models to capture generalized GUI knowledge such as action grounding and task decomposition. However, these benchmarks often overlook structural differences between apps, driven by developer preferences or software architecture, which limits their ability to transfer across unfamiliar environments \cite{deng2024mind2web}. In contrast, interactive environment frameworks \cite{erdogan2024tinyagent, trivedi2024appworld, zhang2024llamatouch, zhou2023webarena} offer real-time learning in realistic environments through state feedback or callable functions. These frameworks provide a “playground" for agents to learn software interaction but face challenges such as high design costs \cite{trivedi2024appworld} and limited transferability across different apps \cite{zhou2023webarena}. Building on insights of these existing benchmarks, we introduce the GUI-Xplore dataset. By capturing exploration videos from apps, GUI-Xplore ensures comprehensive coverage of interaction modes while enhancing scalability through screen recording inputs.

\subsection{Multi-modal GUI Agent}
Recent advancements in large visual-language models (LVLMs), such as BLIP2 \cite{li2023blip}, Qwen2-VL \cite{wang2024qwen2}, and Video-LLaMa \cite{zhang2023video}, have led breakthroughs in multi-modal understanding and reasoning. LVLMs have shown significant promise in a wide range of tasks, including document comprehension \cite{liu2024hrvda, liu2024textmonkey, wan2024omniparser} and GUI understanding \cite{li2024ferret2, you2025ferret}. Recent works have further explored LVLMs' capabilities in GUI automation tasks \cite{zheng2024gpt, hong2024cogagent}. Some approaches focus on enhancing GUI operation capabilities using reinforcement learning \cite{shaw2023pixels, bai2024digirl}, while others emphasize multi-level GUI interpretation through page and HTML fusion \cite{kil2024dual, furuta2023multimodal, gur2023real}. Our work distinguishes itself by focusing on the cross-app and cross-task abilities of GUI agents. Using GUI modeling and the GUI Transition Graph, our baseline model equips the agent to rapidly adapt to new environments.

\section{The GUI-Xplore Dataset}
\label{sec:dataset}
\begin{table*}[htbp]
  \centering
  \begin{tabular}{@{}cccccccccc@{}}
    \toprule
    Method & \multicolumn{3}{c}{Data Format}                                    & Task              & Task      & Sample        & Avg.           & Env.     & Env.  \\ 
    \cmidrule(lr){2-4}
                    & Text   & Image     & Video     &                                              & Num.      &  Num.         & Len.          &        Num.      &   Model    \\ 
    \midrule
    Mind2Web\cite{deng2024mind2web}        & \ding{51} & \ding{51} &            & Task Automation   & 1         & 2350          & 7.3 frame               &   137    &       \\
    AITW\cite{rawles2024androidinthewild}            &           & \ding{51} &  & Task Automation   & 1         & 715142        & 6.5 frame              & 357      &         \\
    WebArena\cite{zhou2023webarena}        & \ding{51} & \ding{51} &            & Task Automation   & 1         & 812           & -                   & 6            & \ding{51} \\
    LLaMATouch\cite{zhang2024llamatouch}      & \ding{51} & \ding{51} &         & Task Automation   & 1         & 495           & 7.01 frame                 &  57    & \ding{51}  \\
    VideoGUI\cite{lin2024videogui}        &           & \ding{51} & \ding{51}   & Task Automation   & 1         & 549           & 55 sec.              & 11           &          \\
        \midrule
    GUI-Xplore            & \ding{51} & \ding{51} & \ding{51}                   & App Exploration  & \textbf{5}     & 32569    & \textbf{21.73 min.}         & 312   & \ding{51} \\
    \bottomrule
  \end{tabular}
  \caption{Comparison of dataset statistics across GUI agent benchmarks. GUI-Xplore provides exploration videos, ensuring data scalability (sufficient Env. Num.) while offering rich prior information crucial for environment modeling(Env. Model). Building on this foundation, GUI-Xplore supports a diverse set of downstream tasks (task num.), delivering a comprehensive dataset to enable the development of versatile and generalizable GUI agents.}
  \label{tab:dataset_comparison}
  \vspace{-2mm} 
\end{table*}

\begin{figure*}[t]
  \centering
  \includegraphics[width=0.9\linewidth]{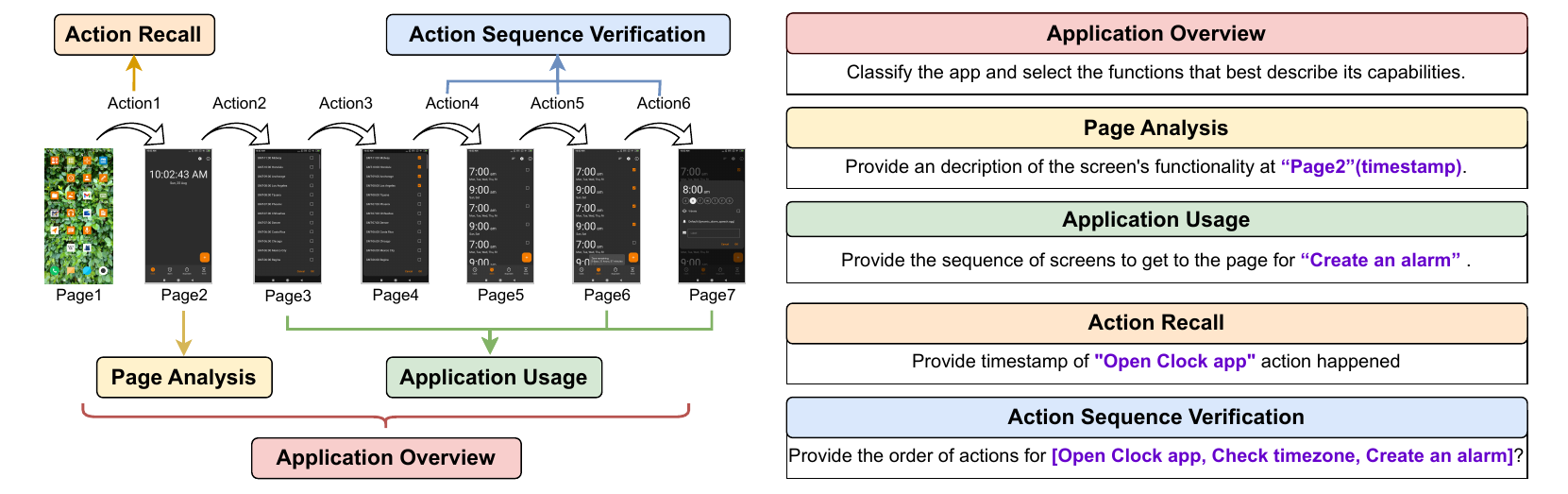}
  \caption{Sample data from five downstream tasks. GUI-Xplore provides app exploration videos paired with five downstream tasks. The videos comprehensively capture all page and action information during the exploration phase. The downstream task employs multiple-choice question answering, targeting different granularity of page and action information. Detailed samples are shown in appendix.}
  \vspace{-5mm}
  \label{fig:dataset}
\end{figure*}

Existing datasets \cite{deng2024mind2web, rawles2024androidinthewild} pay insufficient attention to the diversity in app designs and the varied capabilities required by generalized GUI agents, thereby limiting the performance of existing methods in cross-app and cross-task scenarios. To address these shortcomings, GUI-Xplore introduces two core strategies. First, to facilitate cross-app generalization, we emulate the human strategy of exploration followed by reasoning when encountering unfamiliar apps. Specifically, GUI-Xplore provides models with prior contextual knowledge of each app environment through pre-recorded exploration videos, ensuring a comprehensive coverage of all screens to capture complete structural information. Second, to overcome cross-task limitations, GUI-Xplore defines five hierarchical downstream tasks, focusing on  Environment Understanding and Operational Behavior Understanding, allowing systematic evaluation of model performance across diverse operational contexts.

\subsection{Task Formulation}

The App Exploration task provides the model with exploration videos of unfamiliar apps, requiring the model to quickly learn app-specific knowledge and operational behaviors for effective responses across five downstream tasks. To ensure consistency across tasks with diverse output formats, we standardize all tasks as five-way multiple-choice questions. This standardization facilitates evaluation by aggregating the question-answering accuracy as a unified performance metric.

\subsection{Data Collection}

During the data collection phase, we conduct an in-depth exploration of each app to cover as many interactive pages as possible. This process systematically recorded page information, interaction behaviors, and screen activities. The primary goal of the exploration is to achieve broad coverage, rather than to collect information tailored for specific downstream tasks, resulting in task-agnostic videos. Initially, we utilized an automated exploration tool based on DroidBot \cite{li2017droidbot}, which interprets mobile interfaces and simulates user interactions with clickable elements. This approach enabled automated exploration and annotation across 207 apps. However, recognizing the limitations of automated tools in replicating authentic user behavior, we developed a manual annotation tool built upon Scrcpy\footnote{\url{https://github.com/NetrisTV/ws-scrcpy}.}, an open-source screen mirroring framework. This tool allowed crowd workers to explore apps on cloud-based mobile devices, yielding manual data for 105 popular apps.

Notably, DroidBot\cite{li2017droidbot} leverages the collected View Hierarchy\footnote{Similar to the HTML in website.}(VH) and action data to generate a GUI Transition Graph through a rule-based approach, integrating the discrete information captured during the exploration. This not only provides app-level data for further annotation, but also inspired the development of the Xplore-Agent method.

\subsection{Downstream Task and Annotation}

To develop a versatile GUI agent capable of handling diverse tasks, we designed five downstream tasks focusing on two core aspects: interaction environment understanding and operational behavior analysis. The first category evaluates the model’s understanding of global app functions and specific GUI components, while the second focuses on its ability to grasp temporal and logical relationships within operational sequences. To conduct the annotation, we utilize data collected during the exploration phase, including detailed page information and the GUI transition graph. To provide functional annotations, we utilize AutoDroid’s \cite{wen2024autodroid} Simulated Task Generation method via GPT \cite{achiam2023gpt}, produce functional descriptions and simulated operational tasks for screens. Leveraging these annotations, we create diverse QA pairs with GPT, covering various formats for a comprehensive model evaluation. The example is shown in Fig\ref{fig:dataset}. Moreover, we conducted a manual review of the generated QA pairs to validate their accuracy.

\subsubsection{Environment Understanding}
The Environment Understanding task evaluates the model's capacity to capture both global and local information within an app's interface.

\begin{enumerate}
\item 
\textbf{Application Overview: }The model is tasked with summarizing the core functions of the app based on video data. Annotations are generated using metadata from FDroid\footnote{\url{https://f-droid.org/}.} and subsequently synthesized into detailed software function annotations via GPT.

\item 
\textbf{Page Analysis: }The model is queried to analyze the page function of specific screen. This subtask assesses the model’s ability to extract localized information and functionally interpret GUI interfaces. 

\item 
\textbf{Application Usage: }The model need to infer the operation sequence from homepage to task completion. Specifically, the target sequence is not shown in exploration video, which requires the model to integrate global modeling with local interface understanding.
\end{enumerate}

\subsubsection{Operational Behavior Understanding}

The Operational Behavior Understanding task assesses the model's capacity to comprehend and analyze operational behavior during interactions. 
\begin{enumerate}
\item 
\textbf{Action Recall: }This subtask evaluates the model’s ability to identify the temporal sequence of operational behaviors, necessitating precise localization of specific operations within the video.
\item 
\textbf{Action Sequence Verification:} This subtask examines the model’s capability to establish operation order by assessing the sequence of specified actions based on their global interrelationships. To eliminate operational ambiguity, unique reachability sets for all nodes are generated from the GUI transition graph during data collection. From these sets, operation triples with clear topological relationships are extracted to create comprehensive annotations for this task.

\end{enumerate}

\subsection{Dataset Analysis}

The GUI-Xplore dataset consists of annotations for 312 apps, of which 207 were obtained via automated exploration and 105 through manual annotation. It encompasses 33 subcategories within 6 primary software domains: Entertainment, Productivity, Health, Shopping, Travel, and News. Collectively, the dataset includes 115 hours of exploratory videos, averaging 23.73 minutes per app, establishing comprehensive environmental priors to enhance cross-app generalization in GUI agents. Each app contains annotations for five downstream tasks, totaling 32,569 question-answer pairs to facilitate robust cross-task training and evaluation. To support varied GUI agent approaches, the dataset synchronizes 34,367 visual hierarchy captures, screenshots, and 41,293 action annotations.

Data partitioning was conducted to leverage the deeper traversal and smoother operational flow of manually-collected data. Specifically, 20\% of apps from the manual dataset were allocated to the test set to ensure representation across all six software categories. The remaining 80\% of the manually-collected data, alongside all automated exploration data, formed the training set.

\subsection{Comparison with Existing Dataset and Research Challenges}

The GUI-Xplore dataset represents a framework shift, distinguishing the current pre-training paradigm to exploration-then-reasoning. This framework enhances cross-app adaptability by providing models with app-specific interaction insights via pre-recorded exploratory videos, enabling dynamic adaptation to diverse software environments. Moreover, GUI-Xplore includes five hierarchical downstream tasks that require models to construct a comprehensive, layered understanding of app structures and inter-page relationships, essential for advanced human-device interaction.

This comprehensive design presents novel research challenges for advancing GUI agent development. First, the explore-then-reason framework not only focus on the foundational understanding of GUI components, but also the ability to induce  knowledge for unfamiliar app. Second, the multi-layered downstream tasks require the model to establish global mappings across app structures, progressing beyond single-action generation to attain deeper structural comprehension. Finally, GUI-Xplore’s extensive 20-minute interaction videos, covering approximately 200 pages per app, introduce additional processing complexity yet significantly enrich the model’s contextual understanding.

\section{Baseline Method: Xplore-Agent}

\begin{figure*}[t]
  \centering
  \includegraphics[width=0.9\linewidth]{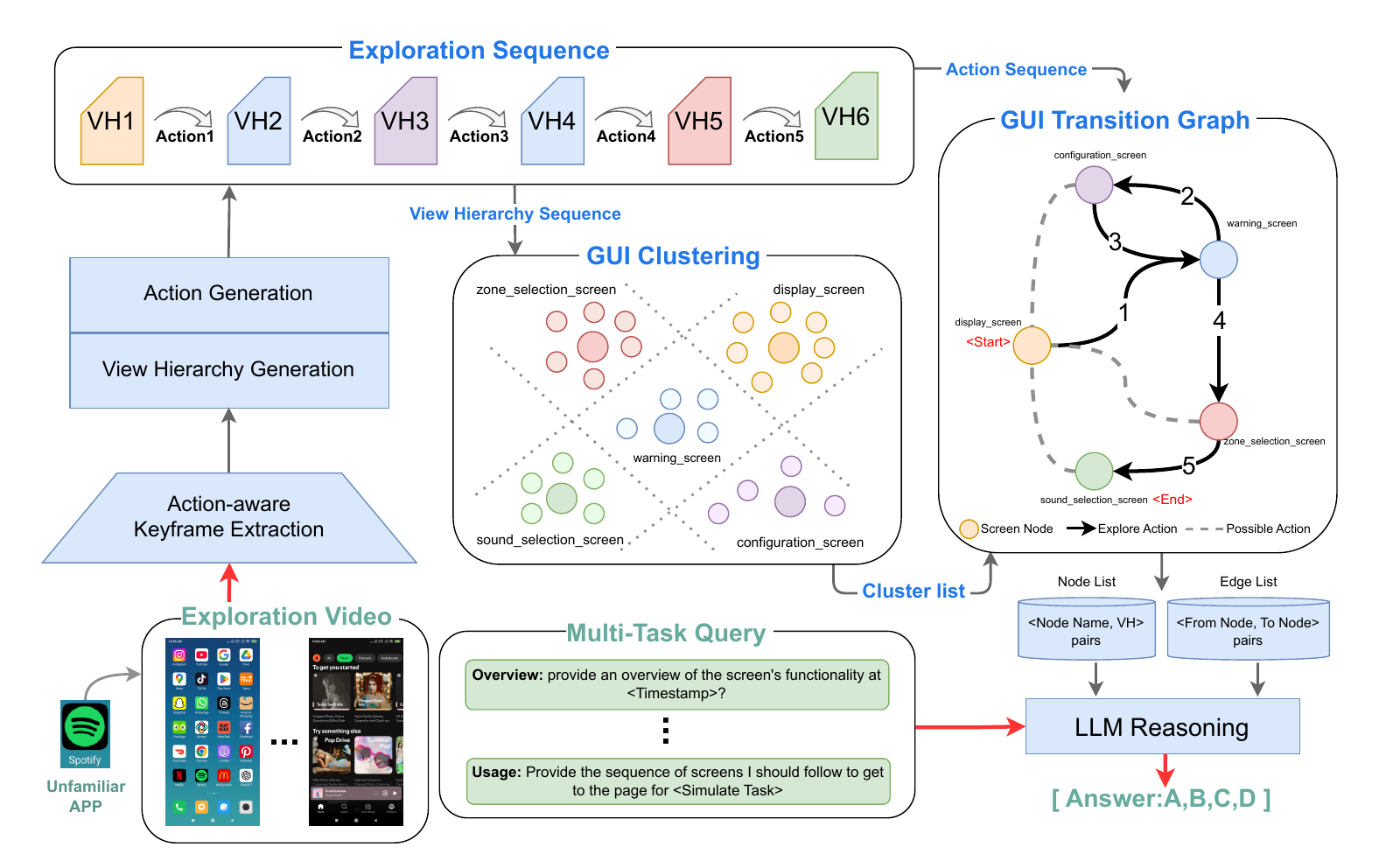}
  \vspace{-2mm} 
  \caption{An overview of the Xplore-Agent pipeline. The model takes an exploration video and a task query as inputs, generating predicted answers. Specifically, the exploration video is converted into a textual exploration sequence through Action-aware Keyframe Extraction, View Hierarchy Generation, and Action Generation. The GUI Clustering Model then groups screens with similar functionalities, transforming the linear sequence into a GUI Transition Graph. Finally, the nodes and edges are used to compose the prompt for querying LLM.}
  \label{fig:method}
  \vspace{-3mm}
\end{figure*}

\label{sec:method}
The App Exploration task in the GUI-Xplore dataset presents challenges due to its information density and the multi-layered nature of the downstream tasks. To address these challenges, we introduce a two-stage framework for GUI agents, as shown in Fig \ref{fig:method}. The first stage employs Action-aware GUI Modeling, wherein local GUI features are extracted from representative keyframes and interaction sequences, enabling the model to capture interactions that are both context-specific and semantically rich. The second stage employs a GUI Transition Graph to model the app’s global environment, integrating the local interactions into a comprehensive representation of the app's structure. Subsequently, this graph-based representation is processed by a LLM to perform downstream Q\&A tasks, allowing the agent to reason effectively across both local interactions and the app's broader structural context.

\subsection{Action-aware GUI Modeling}

Exploration videos offer rich visual and temporal information, but their unstructured format hinders efficient encoding and interpretation. To address this, we model operational behavior by extracting keyframes and generating operational sequences, preserving fine-grained details, such as interaction transitions and state changes.

\subsubsection{Action-aware Frame Extraction}
To capture operational dynamics from the exploration video, we first focus on Action-aware Frame Extraction. Unlike natural video flows, GUI actions often manifest with distinct boundaries, preceded and followed by relatively static states. To accurately detect the beginning and end of actions, we utilize luminance difference (Y-Diff) in the YUV color space \cite{feng2023video2action}, which is sensitive to visual changes and computationally more efficient than alternative metrics, such as structural similarity \cite{wang2004image}. By tracking Y-Diff changes across adjacent frames, key frames marking the start and end of each action are accurately identified.

\subsubsection{Exploration Sequence Generation}
The exploration video encompasses nearly all interactive pages, resulting in an average of 200 keyframes after action-aware extraction. To preserve operational coherence without further reducing the keyframe count, we convert keyframe information into textual representations. This is achieved through VH generation and action generation, enabling data compression while retaining details.

\textbf{VH Generation.} We adopt the pre-training objective of Pix2Struct \cite{lee2023pix2struct} and utilize a Language-Vision Model (LVLM) to generate simplified VH data for each page.

\textbf{Action Generation.} To assist downstream modules in understanding page transformations, we develop a pre-trained action generation module that outputs action categories and associated objects by processing screenshots and simplified VH representations of adjacent frames.

\subsection{Graph-Guided Environment Reasoning}
Unlike the discrete, linear operation flow typical of standard GUI tasks, real-world apps feature intricate and non-linear page transition patterns. To model this complexity, we introduce the GUI Transition Graph to represent the graph structure of the environment in exploration videos, capturing the app's complex page-switching dynamics. Specifically, we cluster similar pages in the video to generate the graph that represent interaction relationships, which guide downstream question-answer reasoning tasks.

\subsubsection{GUI Clustering}

The goal of page clustering is to group pages with shared GUI transition relationships, converting the linear operation flow into a graph. Unlike the data collection phase, where both the real page View Hierarchy (VH) and operational behavior data are available, inference relies solely on the input exploration video and generated intermediate results. To address this limitation, we propose a GUI clustering module based on LVLM. This module leverages the LVLM's ability to understand and summarize local screenshots and operational behaviors while maintaining a list of screen nodes. The model sequentially processes operational keyframes, determining whether the new keyframe belongs to an existing node’s functional category. If they match, the keyframe is grouped with the corresponding node; if not, a new screen node is created, along with a functional description of the page to aid subsequent clustering.
\vspace{-2mm} 
\subsubsection{Graph Generation and Q\&A}

By combining GUI clustering with action information modeling, we generate a GUI Transition Graph. In the graph, nodes represent screen cluster centers, with functional descriptions as node values, while edges connect nodes based on the generated action information. This graph provides a global representation of the interaction environment. Thus we utilize the node list and edge list of the graph to serve as context for downstream Q\&A tasks, guiding the LLM in completing task-specific queries.

\vspace{-2mm} 
\section{Experiments}

\label{sec:expriment}

The GUI-Xplore dataset aims to push the boundaries of cross-app and cross-task capabilities for GUI agents. First, in Section 5.3.1, we demonstrate the improved cross-app generalization performance of Xplore-Agent compared to existing GUI agents. Next, Section 5.3.2 provides comprehensive benchmarking of SOTA models across a range of diverse tasks included in GUI-Xplore, highlighting each model's strengths and limitations. Finally, in Section 5.4, we conduct ablation studies to quantify the impact of each critical component in the Xplore-Agent framework.

\subsection{Experimental setup}
In the Cross-App Generalization experiment, to mitigate the impact of task discrepancies when comparing against existing GUI agent methods, we constructed a \textbf{Cross-App Automation test set} based on the Application Usage task from GUI-Xplore dataset. This test set includes 500 operations spanning 20 diverse software apps and is structured consistently with standard automation benchmarks. All samples were manually curated to \textbf{exclude any overlaps with existing datasets or action sequences that appeared in the exploration videos}, ensuring unbiased evaluation. We compared SOTA GUI agents against our baseline model Xplore-Agent to measure the generalization improvements facilitated by the proposed exploration-based framework.

In the Cross-Task Performance experiment, we evaluated SOTA methods specifically designed for processing long-duration video inputs. This experiment aimed to comprehensively assess the capabilities of current approaches across multiple complex tasks.

\textbf{Baseline Setup. }Xplore-Agent follows a two-stage pipeline consisting of Action-aware GUI Modeling and Graph-guided Reasoning. In the Action-aware GUI Modeling phase, we fine-tuned QwenVL-7B \cite{bai2023qwen} to enhance its understanding of visual and textual features. For the Graph-guided Reasoning phase, GPT was employed to conduct page clustering and contextual reasoning based on the hierarchical structure of the GUI.

\subsection{Competing Methods}

\textbf{Video Understanding Methods.} We categorized the evaluated models into three distinct groups: (1) two-stage models (e.g., VideoTree \cite{wang2024videotree}), (2) open-source end-to-end models (e.g., CogVLM2-Video \cite{hong2024cogvlm2}, VideoChat2 \cite{li2024mvbench}, VideoLLaMA \cite{zhang2023video}), and (3) closed-source end-to-end models (e.g., GPT \cite{achiam2023gpt}). Two-stage models first generate textual features from video frames before leveraging language models for reasoning and question answering, offering improved accuracy in handling long-duration videos at the cost of increased computational complexity. Notably, for models like VideoTree that require textual input, we utilized Pix2Struct to generate captions from GUI content, enhancing their understanding of GUI Screenshots.

\textbf{GUI Agent Methods.} To provide a robust benchmark, we selected three SOTA GUI agent methods for comparative evaluation, including CogAgent\cite{hong2024cogagent}, SeeClick\cite{cheng2024seeclick} and AUTO-UI\cite{zhang2023you}. These methods obtain GUI operation capabilities through extensive training on comprehensive GUI automation datasets.

\subsection{Main Result}
\subsubsection{Cross-App Experiment}

\begin{table}[htbp]
  \centering
  \begin{tabular}{@{}cccc@{}}
    \toprule
    Method                              & Ele. Acc. & Op. Acc.      & StepSR \\ 
    \midrule
    GPT\cite{achiam2023gpt}          & 5.06\%   & 66.12\%        & 4.02\%\\
    AUTO-UI\cite{zhang2023you}          & 7.40\%    & 24.87\%       & 2.17\%\\
    SeeClick\cite{cheng2024seeclick}    & 6.64\%    & --            & 6.64\% \\
    CogAgent\cite{hong2024cogagent}     & 17.18\%   & 73.54\%       & 15.80\% \\
    
    \midrule
    Xplore-Agent                        & 30.73\%  & 84.63\%       & 30.39\% \\
    \bottomrule
  \end{tabular}
  \vspace{-2mm} 
  \caption{Comparison of SOTA GUI agent methods on the Cross-App Task Automation test-set.}
  \label{tab:method_comparison}
  \vspace{-3mm} 
\end{table}

\begin{table*}[!htbp] 
  \centering
  \begin{tabular}{@{}cccccccc@{}} 
    \toprule
    Method        & Frames      & Overview     & Page       & Usage   & Recall   & Seqverify    & Avg.\\ 
    \midrule
    CogVLM2-Video\cite{hong2024cogvlm2} & 24          & 82.00\%       & 56.31\%   & 32.96\%   & 8.03\%    & 6.73\%    & 37.21\% \\
    CogVLM2-Video & 16          & 84.88\%       & 56.75\%   & 48.88\%   & 9.67\%    & 3.85\%    & 40.81\% \\
    CogVLM2-Video & 8           & 84.87\%       & 56.93\%   & 53.56\%   & 5.84\%    & 8.17\%    & 41.87\% \\
    VideoChat2\cite{li2024mvbench}    & 24          & 82.30\%       & 64.42\%   & 67.04\%   & 21.90\%   & 23.87\%   & 51.91\% \\
    VideoChat2    & 16          & 82.00\%       & 66.42\%   & 66.48\%   & 18.98\%   & 18.27\%   & 50.43\% \\
    VideoChat2    & 8           & 86.50\%       & 68.97\%   & 67.04\%   & 18.98\%   & 17.79\%   & 51.86\% \\
    VideoLLaMA2\cite{zhang2023video}   & 16          & 84.00\%       & 69.71\%   & 62.92\%   & 18.43\%   & 17.31\%   & 50.47\% \\
    VideoLLaMA2   & 8           & 86.75\%       & 75.00\%   & 61.24\%   & 21.90\%   & 20.19\%   & 53.02\% \\
    \midrule
    GPT\cite{achiam2023gpt}    & 24          & 96.50\%       & 81.93\%   & 67.04\%   & 21.72\%   & 23.56\%   & 58.15\% \\
    GPT    & 16          & 96.75\%       & 80.47\%   & 66.85\%   & 24.09\%   & 25.96\%   & 58.82\% \\
    GPT    & 8           & 96.88\%       & 82.12\%   & 66.48\%   & 22.6\%    & 28.85\%   & 59.39\% \\
    \midrule
    VideoTree\cite{wang2024videotree}     & 1FPS        & 89.75\%       & 91.05\%   & 65.73\%   & 21.70\%   & 21.61\%   & 57.97\% \\
    Xplore-Agent   & action-aware& \textbf{99.25\%}       & \textbf{92.86\%}   & \textbf{68.21\%}    & \textbf{24.36\%}   & \textbf{36.54\%}   & \textbf{64.24\%} \\
    \bottomrule
  \end{tabular}
  \vspace{-1mm} 
  \caption{Performance comparison of SOTA Video Understanding methods on GUI-Xplore.}
  \label{tab:performance_comparison}
  \vspace{-1mm}
\end{table*}

\begin{table*}[htbp]
  \centering
  \begin{tabular}{@{}cccccccc@{}}
    \toprule
    Cluster     & Overview  & Page      & Usage     & Recall        & SeqVerify     & Avg.          & Avg. Token /App\\ 
    \midrule    
    w/o Cluster & -         & -         & -         & -             & -             & -             & 5144107 \\
    Rule        & 99.62\%   & 92.98\%   & 67.60\%   & 23.62\%       & 25.54\%       & 61.87\%       & 63771\\
    GPT  & 99.25\%   & 92.86\%   & 68.21\%   & 24.36\%       & 36.54\%       & 64.24\%       & 45199\\
    \bottomrule
  \end{tabular}
  \vspace{-2mm} 
  \caption{Comparison of different GUI clusters methods.}
  \label{tab:gui_cluster_comparison}
  \vspace{-4mm}
\end{table*}

In Table \ref{tab:method_comparison}, we present the performance of existing SOTA GUI agents alongside our proposed baseline, Xplore-Agent, evaluated on the Cross-App test set. Following the evaluation metrics of the Mind2web benchmark, we include Element Accuracy (Ele. Acc.), Operation Accuracy (Op. Acc.), and Step Success Rate (StepSR).

Without automation-specific pre-training, Xplore-Agent effectively integrates general knowledge from LVLMs with app-specific knowledge acquired during the exploration phase. This combination facilitates robust knowledge transfer in unfamiliar software environments, demonstrating a notable 10\% increase in all three metrics compared to existing SOTA methods.

\subsubsection{Cross-Task Experiment}

In Table \ref{tab:performance_comparison}, we present the performance of SOTA long video comprehension methods alongside our proposed Xplore-Agent across five downstream GUI  tasks in GUI-Xplore.

\textbf{Comparison of Methods: }The experimental results demonstrate that two-stage video comprehension methods tend to outperform end-to-end approaches across various tasks. This superior performance can be attributed to their denser frame sampling and the efficient compression of extensive information using textual descriptions, effectively leveraging the exploration video data. Xplore-Agent effectively models global app information using a GUI Transition Graph, resulting in a significant performance improvement compared to the two-stage method  VideoTree.  Additionally, GPT surpasses open-source end-to-end models in overall GUI agent capabilities, but incurs significantly higher token processing costs during video input.

\textbf{Task Performance Analysis:} The results reveal a substantial performance disparity across the five types of GUI navigation tasks. All models show superior performance on Environment Understanding tasks compared to Operational Behavior Understanding tasks. This suggests that existing LVLMs possess inherent GUI comprehension capabilities acquired from large-scale pre-training, yet there remains considerable room for improvement in understanding inter-page relationships and GUI interaction behaviors. Notably, Xplore-Agent also demonstrates performance gains on those challenging tasks.

\textbf{Impact of Input Frame Count: }We analyzed how varying input frame counts affect end-to-end model performance. Notably, increasing the frame count often degraded performance, indicating that more input frames do not necessarily improve environment understanding. Instead, excessive input data can overwhelm the model, increasing ambiguity and reducing performance.

\subsection{Ablation Experiment}

We conduct ablation experiments to investigate the influence of each component and design in GUI-Xplore.

\subsubsection{GUI clustering}

Xplore-Agent encodes GUI modeling-generated linear sequences into a GUI Transition Graph through GUI clustering, where clustering accuracy is critical for model performance. We performed ablation studies to compare our clustering method against a traditional rule-based approach, which utilizes View Hierarchy (VH) and screenshot similarity for clustering. As shown in Table \ref{tab:gui_cluster_comparison}, our clustering method effectively reduces token usage while simultaneously improving prediction accuracy, highlighting its advantages in GUI representation and task inference.

\subsubsection{Action-aware keyframe extraction}
\begin{table}[htbp]
  \centering
  \begin{tabular}{@{}ccc@{}}
    \toprule
    Keyframe Method         & Num./app & Page Num./100 frames \\ 
    \midrule
    1 FPS         & 629.35   & 54.95                \\
    Action-aware  & 115.75   & 10.11                \\
    \bottomrule
  \end{tabular}
  \vspace{-2mm} 
  \caption{Comparison of keyframe extraction methods.}
  \label{tab:keyframe_comparison}
  \vspace{-3mm}
\end{table}
Due to the substantial information contained in exploration videos, it is essential to minimize the number of keyframes. While fixed-interval sampling is a common strategy, it overlooks the distinctive interaction patterns in GUI videos. As illustrated in Table \ref{tab:keyframe_comparison}, the action-aware extraction approach effectively eliminates redundant frames, thereby reducing the computational burden for subsequent analysis.

\vspace{-2mm} 
\section{Conclusion}
\label{sec:conclusion}
To tackle the generalization gap in current GUI agents across diverse apps and tasks, we present GUI-Xplore. GUI-Xplore provides exploration videos as rich priors for unfamiliar software, boosting performance in varied tasks. Based on this, we introduce Xplore-Agent, a baseline framework utilizing a GUI Transition Graph to model the exploration environment. Experiments show that this approach significantly improves cross-app generalization and multi-task capabilities. We anticipate that the exploration-based framework will advance the development of more versatile and generalizable GUI agents.

\textbf{Limitation.} To enhance task generality, GUI-Xplore only requires the model to output textual answers instead of concrete actions, indicating that our work has not yet achieved a fully generalized GUI agent. Additionally, the collection of exploration videos faces challenges related to data sources and privacy concerns. These limitations are discussed in detail in the appendix.

\section*{Acknowledgements}
This work is supported by National Natural Science Foundation of China (Grant No.62272261), Tsinghua University (AIR)–AsiaInfo Technologies (China) Inc. Joint Research Center, Wuxi Research Institute of Applied Technologies, Tsinghua University and Beijing Academy of Artificial Intelligence (BAAI).

This work was partly funded by the Shanghai Municipal Science and Technology Major Project (2021SHZDZX0102), and STCSM (22DZ2229005).

{
    \small
    \bibliographystyle{ieeenat_fullname}
    \bibliography{main}
}

\end{document}